\documentclass{article}

\usepackage{arxiv}

\usepackage[utf8]{inputenc} 
\usepackage[T1]{fontenc}    
\usepackage{hyperref}       
\usepackage{url}            
\usepackage{booktabs}       
\usepackage{amsfonts}       
\usepackage{nicefrac}       
\usepackage{microtype}      
\usepackage{graphicx}
\usepackage{natbib}
\usepackage{doi}
\usepackage{algorithm}
\usepackage{amsmath}
\usepackage{bm}
\usepackage{multirow}
\usepackage{algpseudocode}
\usepackage{float} 
\usepackage{amsmath,amssymb} 

\newcommand\blfootnote[1]{}
\usepackage{subcaption} 

\newcommand{\argmax}{\mathop{\mathrm{argmax}}\limits} 

\title{A Competitive Learning Approach for Specialized Models: A Solution for Complex Physical Systems with Distinct Functional Regimes}

\author{ \href{https://orcid.org/0000-0003-2865-3997}{\includegraphics[scale=0.06]{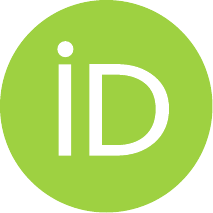}\hspace{1mm}Okezzi F.~Ukorigho}\\
	Department of Mechanical Engineering\\
	Louisiana State University\\
	Baton Rouge, LA 70803 \\
	\texttt{oukori1@lsu.edu} \\
	\And
	\href{https://orcid.org/0000-0001-5531-2245}{\includegraphics[scale=0.06]{orcid.pdf}\hspace{1mm}Opeoluwa ~Owoyele} \\
	Department of Mechanical Engineering\\
	Louisiana State University\\
	Baton Rouge, LA 70803 \\
	\texttt{oowoyele@lsu.edu} \\
}

\renewcommand{\shorttitle}

\hypersetup{
pdftitle={A template for the arxiv style},
pdfsubject={q-bio.NC, q-bio.QM},
pdfauthor={David S.~Hippocampus, Elias D.~Striatum},
pdfkeywords={First keyword, Second keyword, More},
}

\begin{document}
\maketitle

\begin{abstract}
Complex systems in science and engineering sometimes exhibit behavior that changes across different regimes. Traditional global models struggle to capture the full range of this complex behavior, limiting their ability to accurately represent the system. In response to this challenge, we propose a novel competitive learning approach for obtaining data-driven models of physical systems. The primary idea behind the proposed approach is to employ dynamic loss functions for a set of models that are trained concurrently on the data. Each model competes for each observation during training, allowing for the identification of distinct functional regimes within the dataset. To demonstrate the effectiveness of the learning approach, we coupled it with various regression methods that employ gradient-based optimizers for training. The proposed approach was tested on various problems involving model discovery and function approximation, demonstrating its ability to successfully identify functional regimes, discover true governing equations, and reduce test errors.
\end{abstract}


\section{Introduction}
\par In the era of data-driven science, machine learning has emerged as a transformative tool, offering unprecedented solutions to complex problems across a wide range of scientific and technological domains. Specifically, machine learning has found applications in diverse fields such as biology, medicine, material science, engineering, energy, manufacturing, and agriculture. Notable examples include rapid detection of SARS-CoV-2 \cite{ikponmwoba2022machine}, advances in drug discovery and development \cite{talevi2020machine}, quality control and defect detection \cite{wang2022process}, climate modeling and prediction \cite{krasnopolsky2006complex}, as well as crop yield forecasting and optimization \cite{di2022new}.

Some of these applications involve classification, which involves learning based on categorical data. In this regard, machine learning techniques, such as Support Vector Machines, Naive Bayes, K-nearest neighbor, and Neural Networks, have been used to extract texture features from images for subsequent classification \cite{chola2022bcnet, chola2022gender}. Additionally, machine learning has facilitated the identification of high-order closure terms from fully kinetic simulations, a critical aspect of multi-scale modeling\cite{laperre2022identification}. On the other hand, function approximation or regression involves estimating a continuous target quantity as a function of a set of input variables. Methods such as Sparse Identification of Nonlinear Dynamics (SINDy) \cite{brunton2016discovering}, the Least Absolute Shrinkage and Selection Operator (LASSO) \cite{tibshirani1996regression}, Dynamic Mode Decomposition (DMD) \cite{schmid2010dynamic,mezic2005spectral}, Koopman operator \cite{mezic2013analysis} and the Eigensystem Realization Algorithm (ERA) \cite{juang1985eigensystem} have contributed significantly to understanding complex systems by offering effective strategies for model selection, variable regularization, decomposition of high-dimensional systems, and extraction of state-space models from input-output data.

However, many systems exhibit localized behaviors, characterized by changes in the physical characteristics or governing laws that vary based on time or space, input variables, or system parameters. For instance, in non-reacting fluid mechanics, the physical behavior of a viscous fluid depends on the Reynolds number (Re) \cite{li2015fsi}, with low Re corresponding to viscous-dominated laminar flow, and high Re indicating chaotic convection-dominated turbulent flows. This has implications for empirical correlations of flow resistance through pipes and drag force estimations of immersed bodies; such correlations are often piecewise and split based on the value of Re. A similar phenomenon exists in turbulent reacting flows; various burning zones and turbulence-chemistry regimes exist (e.g., broken reaction, corrugated flamelets) based on the characteristic chemical and flow time scales. In the context of dynamical systems modeling, specialized models can be employed to represent different regimes in systems such as tumor growth models, where distinct regimes can be used to capture the time-dependent response to various treatment methods \cite{atangana2022piecewise}. Similar situations arise in various domains, including photovoltaic grid-connected systems \cite{li2020piecewise}, legged locomotion \cite{holmes2006dynamics}, and mass-spring-damper systems \cite{dai1994oscillatory}. In some cases, changes can be described via simple piecewise functions, while in other cases, the boundaries between various regimes may be fuzzy and marked by high-dimensional demarcations in vector space. Model identification for such datasets presents a unique challenge due to the difficulty in accurately modeling and segregating these distinct domains \cite{mangan}.

To learn functions involving these kinds of systems, specialized models can be employed. This approach involves training multiple models, each specializing in making predictions within specific zones based on input variables, physical space or time, or model parameters.
In cases where machine learning models are used to approximate such systems, it is desirable to split the learning problem amongst specialized models. As a result, multiple models can be trained: each being the expert in making predictions at specific zones, depending on the input variables, physical space or time, or the model parameters. This approach to learning, therefore, first identifies distinctions between different regimes of the problem, then before learning models that apply locally in each regime. Such an approach has a number of advantages over the conventional approach of using a global model. First, such specialized models tend to have lower complexity and higher accuracy since they focus on localized regions instead of attempting to learn the entire dataset comprising various zones. First, such models, since they are specialized to localized regions of the problem, will lead to models with lower complexity and higher accuracy, compared to a global model which attempts to learn the entire dataset consisting of various zones. Moreover, such specialized models, when successful in discovering the true regimes, result in models that are more consistent with the underlying physics, thereby enhancing the potential for improved generalization. In contrast, approaches that involve global machine learning models are unable to adequately capture the behavior of these systems, due to their inability to segregate and model distinct regimes of functions.

One of the first attempts to extract specialized models from data was the mix of experts (MoE) approach, introduced by Jacobs et al. \cite{jacobs1991adaptive}. This approach devised a loss function that was dependent on the relative performance of each expert, assigning a larger learning rate to competent models, and conversely reducing the learning rate of poorer models. The MoE approach and its variants have found successful applications in various fields such as image recognition \cite{armi2021texture}, natural language processing \cite{du2022glam}, time series prediction \cite{milidiu1999time}, and turbulent combustion modeling \cite{owoyele2021efficient}. In the context of dynamical system identification, other approaches have been developed. For example, Hybrid-SINDy extends the concepts of SINDy to hybrid dynamical systems, identifying the underlying equations without prior knowledge of regime boundaries and leveraging information theory to manage the uncertainty of model selection \cite{mangan}. Another technique utilizes Fourier series decomposition to analyze piecewise nonlinear dynamical systems, detecting piecewise features and representing each segment using a pure Fourier series \cite{wang2023identification}. Other approaches have also been proposed, based on variational expectation maximization \cite{alameda2021variational} and neural fuzzy networks \cite{liu2019identification} for piecewise linear systems.

In this paper, we present a competitive learning approach for specialized models (CLSM), where each model learns to make predictions in localized regimes, rather than being trained globally. In contrast to some of the aforementioned approaches, the proposed approach can be applied to a wide range of problems, ranging from function approximation trained on large datasets using deep neural networks, to dynamical systems identification using LASSO regression. Our approach involves training a collection of models concurrently on a dataset through a competitive learning process that divides the problem into regimes, with each model specializing in making predictions within its assigned regime. In the subsequent sections of this paper, we provide mathematical and visual descriptions of our approach and present validation studies using a synthetic 1-dimensional test problem, damped harmic oscillators, and a combustion data set that involves flame speed predictions. Finally, the paper concludes with remarks and directions for future extensions of this work.

\section{Background and Methodology}
We consider the generic learning problem, where the objective is to learn a model that describes a dataset $\mathcal{D} = {\bm{x}i, y_i}{i=1}^{S}$, where $S$ is the number of observations of the physical system, collected from experiments or numerical simulations. Each $\bm{x}_i \in \mathbb{R}^{n_v \times 1}$ denotes the $i$-th observation of the input variables, while $y_i$ corresponds to the continuous response or observed output for that particular observation. In general, we seek to learn the parameters $\Theta$ of a model, $\mathcal{M}$, in order to minimize a loss function that incorporates a measure of prediction errors.

One commonly used model for estimating ${y}_i$ involves utilizing fully connected artificial neural networks, sometimes simply referred to as neural networks (NNs). These models consist of successive layers that apply linear transformations and nonlinear activations to the preceding layer. The first hidden layer can be represented as:

\begin{equation}\label{Eq:NN_first}
\mathcal{H}^1 = \sigma \left(\bm{W}_1 \bm{x} + \bm b_1 \right)
\end{equation}
and every successive hidden layer as:
\begin{equation}\label{Eq:NN_hidden}
\mathcal{H}^{i} = \sigma \left(\bm W_i \mathcal{H}^{i-1} + \bm b_i \right) 
\end{equation}

Consequently, the final prediction of the neural network can be described through a composition operator $\circ$, as follows:
\begin{equation}\label{Eq:NN_composition}
\hat{{y}} = \mathcal{H}^{n_l} \circ \mathcal{H}^{n_l-1} \circ \; ... \; \circ \mathcal{H}^{1} \left(\bm{x} \right)
\end{equation}

The neural network, as described in Eq. \ref{Eq:NN_first} -- \ref{Eq:NN_composition} consists of an input layer, $n_l$ hidden layers, and an output layer. ${\Theta} = \{ \bm W_i, \bm b_i\}_{i=1}^{n_l}$ is a set containing the trainable parameters, with $\bm W_i$ representing the weight matrix of the $i$-th layer and $\bm b_i$ containing the biases. $\sigma$ is an activation function that enhances the neural network's ability to learn non-linear function approximations. During training, neural networks are trained using gradient-based optimizers of first- or second-order (e.g., stochastic gradient descent \cite{amari1993backpropagation}, Adam optimizer \cite{kingma2014adam}, or Newton optimization). For a single-variable target, the loss function is often the mean squared error, described as

\begin{equation}\label{Eq:loss_function}
\mathcal{L}^{\textrm{MSE}} = \frac{1}{S}\sum\limits_{i=1}^{S} \left(y_i - \hat{y}_i \right)^2
\end{equation}

On the other hand, LASSO is a regression approach that simultaneously conducts a variable selection and regularization \cite{tibshirani1996regression}. LASSO regression is particularly beneficial in high-dimensional settings with redundant features, as it promotes model parsimony and reduces overfitting. LASSO employs the $\ell_1$-norm of the regression coefficients as a penalty on the loss function, thus shrinking some coefficients toward zero. This property enables LASSO to automatically select relevant features and effectively reduce the number of variables in the model. As a linear regression model, LASSO predicts the target variable as:

\begin{equation}\label{Eq:loss_function}
\hat{y}_ = \beta_0 + \sum\limits_{i=1}^{P} \beta_i x_{ij},
\end{equation}

where $x_{ij}$ represents the $j$-th variable of the $i$-th observation, and $\bm{\beta} = \left[\beta_1, \beta_2, ..., \beta_{P-1} \right]$ are the model coefficients. The loss function in LASSO is defined as: 

\begin{equation}\label{Eq:LASSO_loss}
\mathcal{L}^{\textrm{LASSO}} = \mathcal{L}^{\textrm{MSE}} + \mathcal{L}^{\textrm{reg}} = \frac{1}{S}\sum\limits_{i=1}^{S} \left(y_i - \hat{y}_i \right)^2 + \lambda \sum\limits_{j=1}^{P} \lvert \beta_j \rvert
\end{equation}

where $\lambda$ is a scalar that controls regularization. and the second term on the right-hand side represents the penalty term, encouraging the optimizer to find sparse representations of $\bm \beta$. The second term on the right-hand side of Equation \ref{Eq:LASSO_loss} represents the penalty term, which encourages the optimizer to find sparse representations of the coefficient vector $\bm{\beta}$. By adjusting the value of $\lambda$, we can control the trade-off between the goodness of fit and the sparsity of the model. In this study, both NNs and LASSO regression are used to obtain specialized models for the application problems discussed in later sections.

\subsection{Mean squared error loss as a weighted sum of individual errors}
In the previous section, it was highlighted that machine learning models aim to find the optimal values of the parameters, denoted as ${\Theta}$, by minimizing the loss, oftentimes defined as some function of the error.

\begin{figure}[h!]
    \centering
    \begin{subfigure}{0.45\textwidth}
        \centering
        \includegraphics[width=\linewidth]{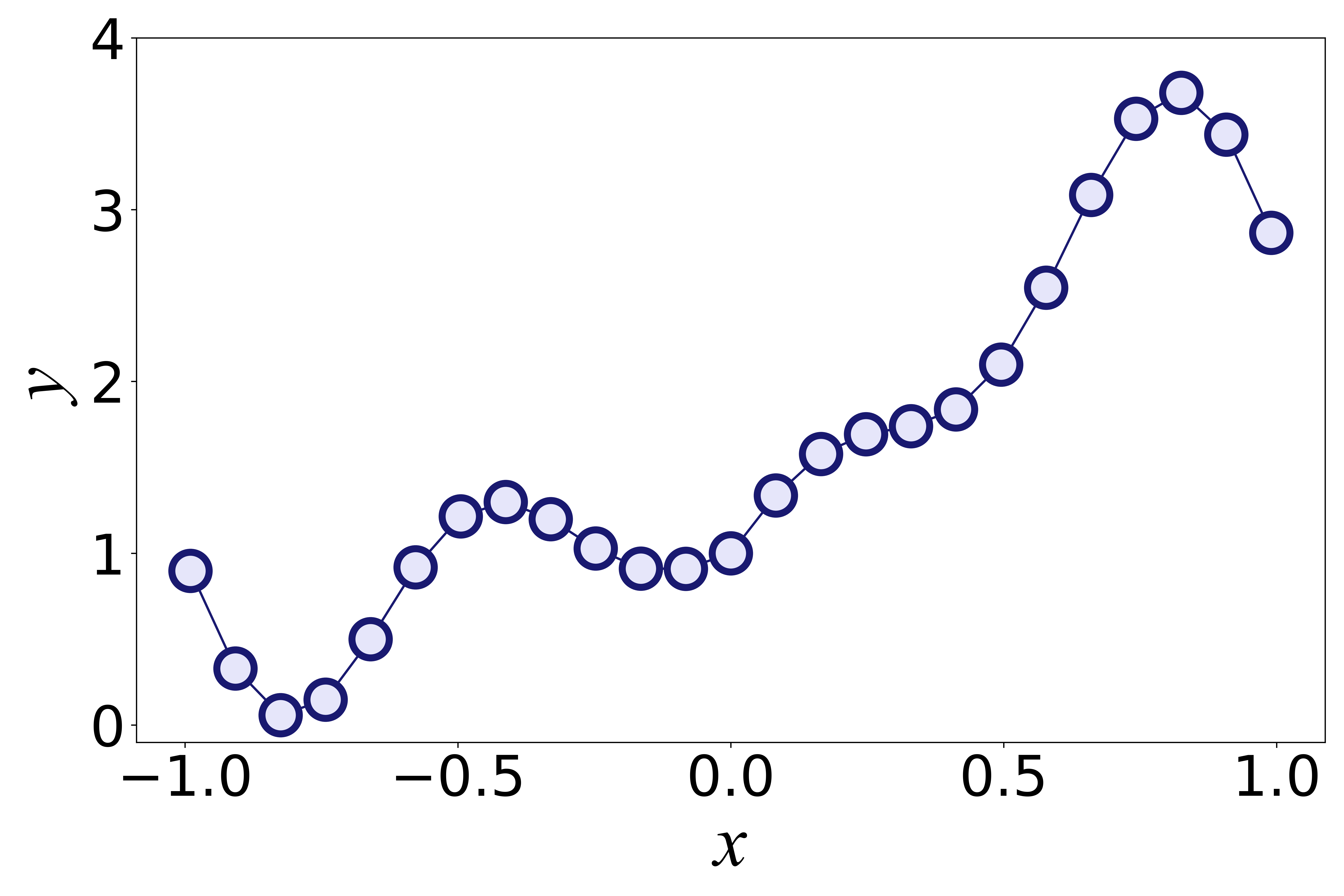}
        \caption{$y$ against $x$}
        \label{fig:toy_truefun}
    \end{subfigure}
    \hfill
    \begin{subfigure}{0.45\textwidth}
        \centering
        \includegraphics[width=\linewidth]{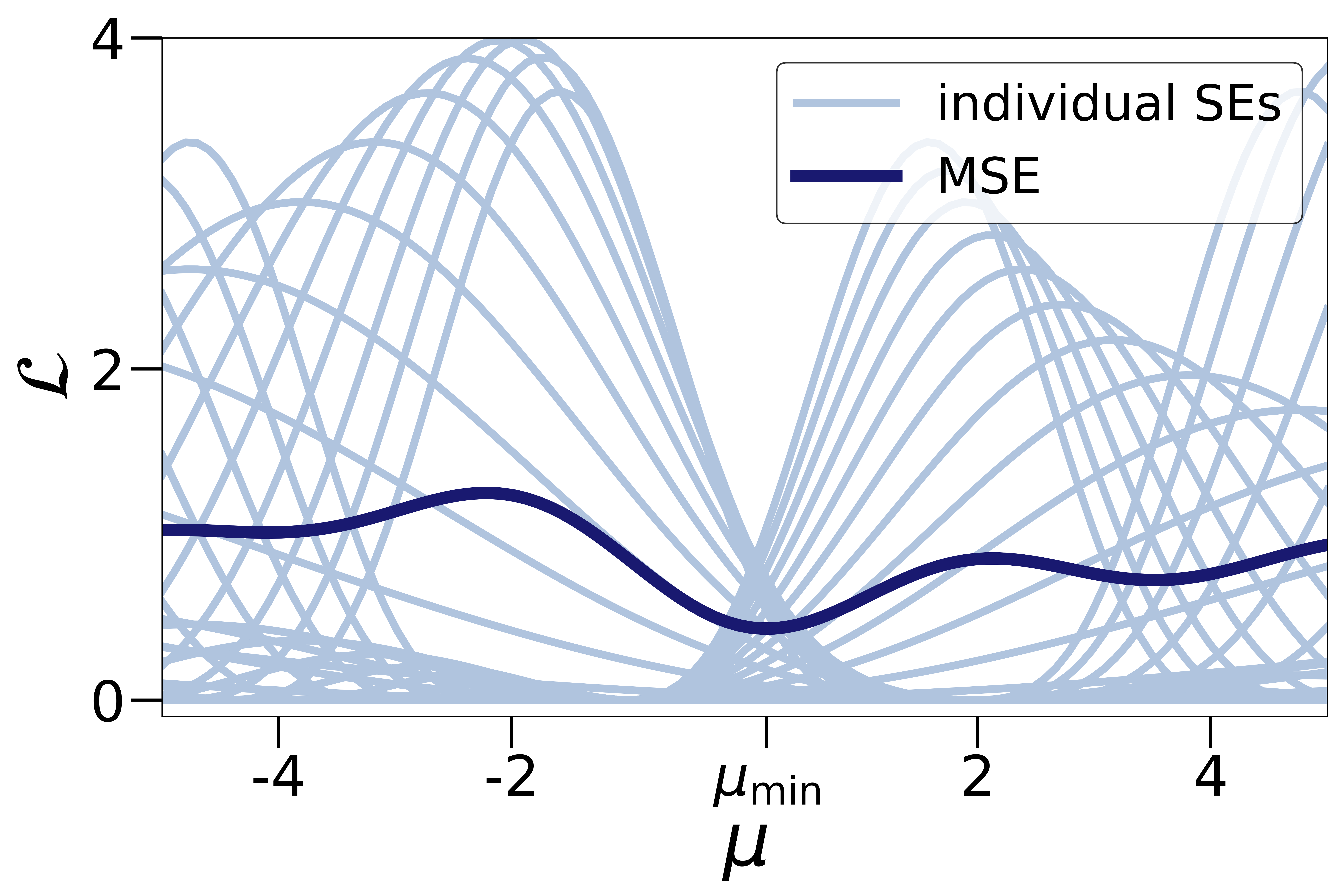}
        \caption{loss landscape for all training samples}
        \label{fig:losslandscape_all}
    \end{subfigure}
    
    \begin{subfigure}{0.45\textwidth}
        \centering
        \includegraphics[width=\linewidth]{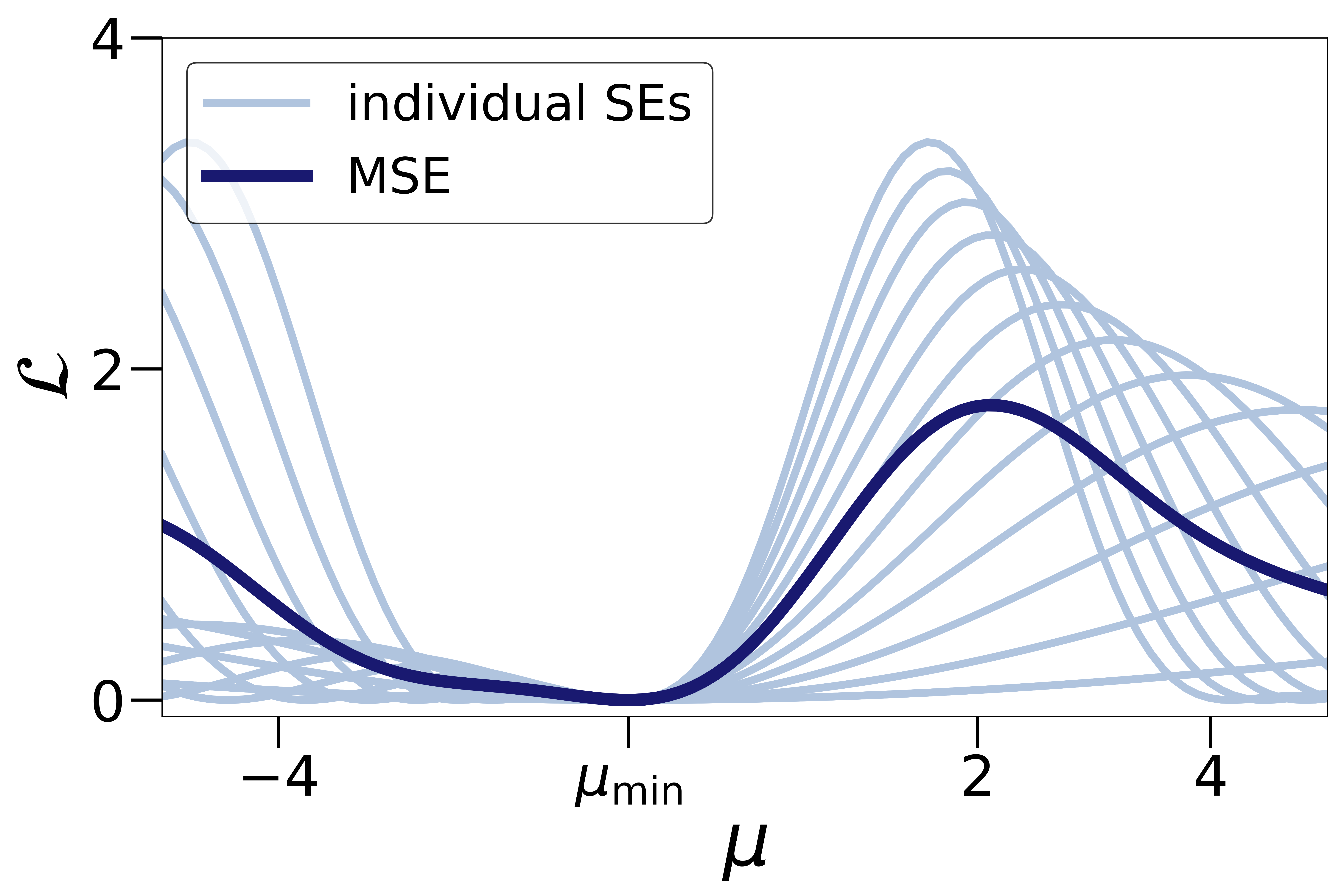}
        \caption{loss landscape for points where $x < 0$}
        \label{fig:losslandscape1}
    \end{subfigure}
    \hfill
    \begin{subfigure}{0.45\textwidth}
        \centering
        \includegraphics[width=\linewidth]{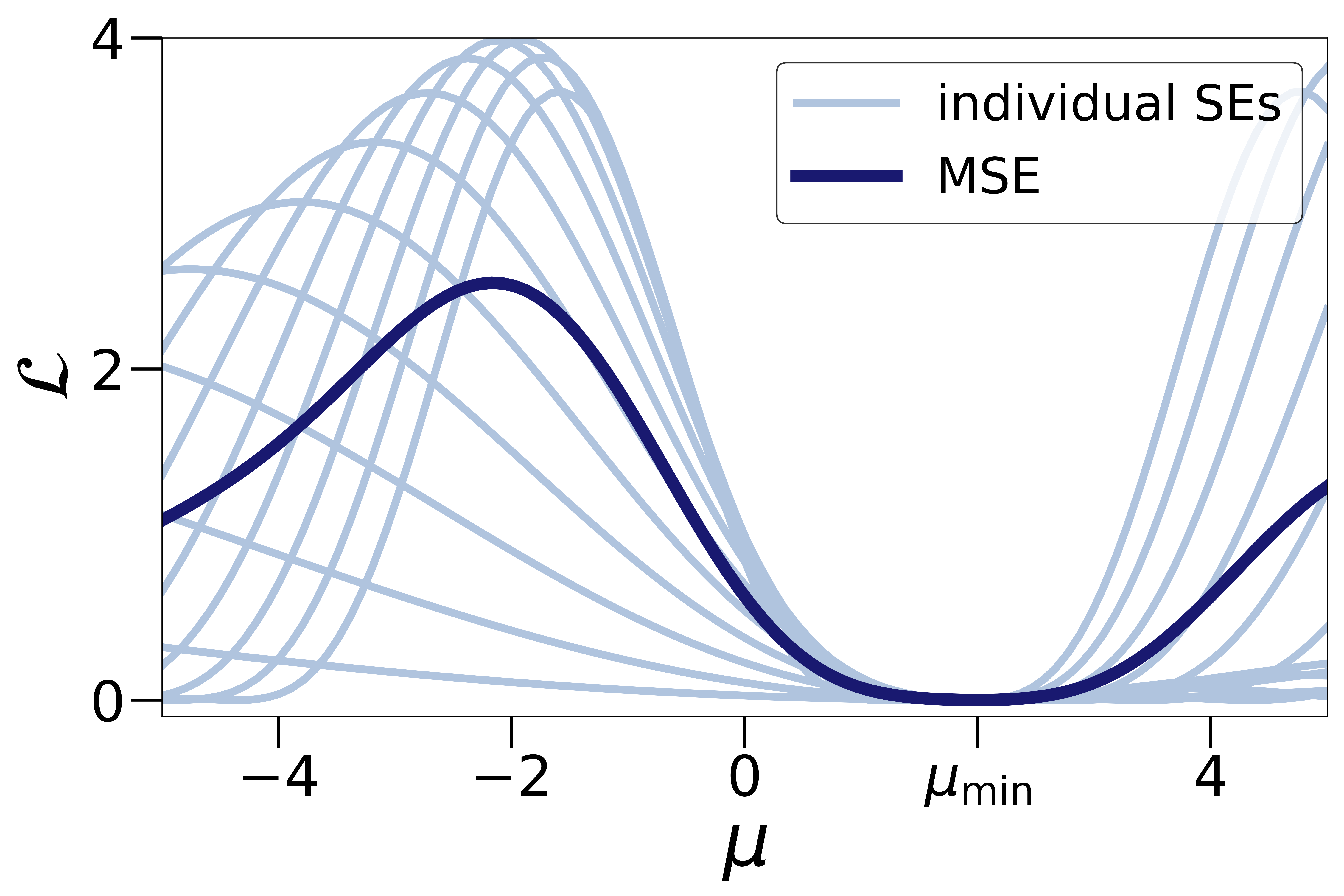}
        \caption{loss landscape for points where $x \geq 0$}
        \label{fig:losslandscape2}
    \end{subfigure}
    
    \caption{Plot of $y$ vs $x$ and variation of loss landscape with selected observations.}
    \label{fig:losslandscape}
\end{figure}

One way to interpret this minimization problem is to consider that the optimizer minimizes a weighted sum of errors for each observation, where each point measurement has a weight of $1/S$. To illustrate this, we consider a synthetic problem shown in Fig. \ref{fig:toy_truefun}, where the true function is given by:

\begin{equation}\label{Eq:toyprob1}
y = f(x) = \sin(\mu x) + x\cos(8 x) + e^{0.1 x} + 1.2 \tanh(x)
\end{equation}
with
\begin{equation}\label{Eq:toyprob2}
\mu =
\begin{cases}
-1, & \text{for } t < 0 \\
2, & \text{otherwise}
\end{cases}
\end{equation}

Here, we assume that we need to fit a model that minimizes the mean squared error (MSE) loss between the actual and model-predicted values of $y$. For illustration purposes, we assume that the functional form of the model and the values of all coefficients and parameters are known, except the value of $\mu$. Our goal is to find the optimal value of $\mu$ that closely approximates the observed response.

As described earlier, a gradient-based optimizer can be used to find values of $\mu$ that lead to predictions that closely approximate the observed response. Figure \ref{fig:losslandscape_all} shows the loss landscape, which represents the MSE as a function of $\mu$. In the plot, the light blue transparent lines depict the losses of the individual observations, denoted as $l_i = \left(y_i - \hat{y}_i\right)^2$, while the solid line represents the overall MSE. In this way, the MSE loss minimized can be interpreted as the overall impact of the squared error losses of all the individual observations. In other words, the final value of $\mu$ is the value of $\mu$ that minimizes the $l_i$'s, on average, despite the fact that the minimum and maximum values of $\mu$ for individual observations may differ.

A careful inspection of Fig. \ref{fig:losslandscape_all} shows that the individual minima of some individual loss functions form clusters. This is more clearly seen in Figs. \ref{fig:losslandscape1} and \ref{fig:losslandscape2}, where only a subset of points in the observations are considered. Specifically, when we focus on data points where $x < 0$, the mean loss landscape is altered, resulting in a global minimum at $\mu = -1$. Similarly, if we only consider data points where $x \geq 0$, the MSE loss is minimized at $\mu = 2$. This observation suggests that training samples generated using similar parameters tend to have individual loss landscapes that converge near the global minima. Overall, it is seen that the loss landscape of an optimizer can be influenced by the observations being considered.
Based on these insights, it is more appropriate to approach the synthetic problem as an optimization problem for two locally appropriate models: one for $x < 0$ and another for $x \geq 0$, each with its own unique loss function. However, in practical scenarios, such splits are not known \textit{a priori}. Therefore, the proposed approach, CLSP, provides a methodology for simultaneously partitioning and learning locally valid models by employing a competitive learning process. In the following section, specific details regarding the working principles of CLSP will be presented.

\subsection{Extracting specialized models from data}\label{sec:clsm}
In this section, we describe the CLSM approach for extracting a set of specialized models from data. As opposed to training a single model, we concurrently train \(Q\) specialized models, denoted by \(\mathcal{M} = \{M_1, M_2, ..., M_Q\}\), on the dataset. We extend the MSE loss in Eqs. \ref{Eq:loss_function} and \ref{Eq:LASSO_loss} by introducing a weight for each sample, resulting in the following modified MSE loss:

\begin{equation}\label{Eq:loss_function_local}
\bar{\mathcal{L}}_k^\textrm{MSE} = \frac{1}{S}\sum\limits_{i=1}^{S} \alpha_{i,k} \left(y_i - \hat{y}_{i,k} \right)^2,
\end{equation}

where $\alpha_{i,k}$ represents the weight of observation $i$ for the $k$-th local model. During training, the models learn through a competitive process, where \(\alpha_{i,k}\) is computed based on the performance of each model relative to others. The goal is to compute \(\alpha_{i,j}\) such that the effects of some observations are amplified, while others are diminished. This can be interpreted as providing a measure of the probability that model \(k\) is the best at predicting \(y_i\), given the input, \(x_i\). The computation of \(\alpha_{i,k}\) is defined as follows:

\begin{equation}\label{eq:model_weight}
\alpha_{i,k} = \frac {\exp \left(-\kappa \left(y_i - \hat{y}_{i,k} \right)^2 /c_i \right)}{\sum\limits_{j=1}^{Q} \exp \left(-\kappa \left(y_i - \hat{y}_{i,k} \right)^2 /c_i\right)}.
\end{equation}

where $c_i$ represents the squared error (SE) of the best model for a given observation, expressed mathematically as $c_i = \min\limits_{k \in Q} \left\{\left(y_i - \hat{y}_{i,k} \right)^2 \right\}_{k=1}^Q$. The parameter $\kappa$ determines the extent of separation between the models, with higher values of $\kappa$ resulting in greater distinctness among the models. It is important to note that as a consequence of the expression for $\alpha_{i,k}$ in Eq. \ref{eq:model_weight}, $0 \leq \alpha_{i,k} \leq 1$ and $\sum\limits_{k} \alpha_{i,k} = 1$ for all $i = 1,2,...,S$. 

During the training process, each model competes with others, “winning” data points that they predict more accurately than the rest. At the same time, the best model for a given observation also negatively alters the loss landscape of other specialized models to prevent them from winning similar observations in future training iterations. Essentially, in the computation of the Mean Squared Error (MSE) loss, if a specific model performs better than others for an observation, that observation is assigned a weight close to 1 in the MSE calculation, while the weights assigned to the other models for that observation are close to zero. Consequently, each model learns not only based on its individual loss landscape but also influences the training of other models by adjusting the weights assigned to their respective observations. This iterative process leads to increasing specialization of models towards specific regions of the input or parameter space throughout training.

In initial tests, the authors observed that gradient-based optimization methods tend to find only locally optimal points when using the MSE loss described in Eq. \ref{Eq:loss_function_local}. This issue arises when a specialized model captures a small cluster of points that should actually belong to another model, leading to a permanent alteration of the loss landscape for the losing model. Consequently, it becomes challenging for the appropriate model to regain those mislabeled points. To overcome this limitation, two potential approaches are proposed.

The first option involves utilizing a global optimizer to perform a comprehensive search, and then using the resulting solution as an initial point for subsequent gradient-based optimization. While this strategy is effective for problems with a small number of trainable parameters, it may not be suitable for more complex tasks involving moderately deep neural networks for function approximation endeavors.

Alternatively, the authors suggest further modifying the loss landscape to alleviate the impact of locally mislabeled points. To accomplish this, a smoothed version of the weight parameter $\alpha$ is computed, denoted as $\bar{\alpha}_{i,k}$. Specifically, $\bar{\alpha}_{i,k}$ is calculated by averaging the weights of its $N$ nearest points: $\bar{\alpha}_{i,k} = \frac{1}{N}\sum\limits_{j \in \mathcal{G}} \alpha_{j,k}$, where $\mathcal{G}$ represents the set of the indices of these $N$ nearest points, determined using the Euclidean distance as a proximity measure.

Subsequently, the final MSE loss used for the optimization in the CLSM approach is given as: 

\begin{equation}\label{Eq:loss_function_final}
\hat{\mathcal{L}}_k^\textrm{MSE} = \frac{1}{S}\sum\limits_{i=1}^{S} \hat{\alpha}_{i,k} \left(y_i - \hat{y}_{i,k} \right)^2,
\end{equation}

where $\hat{\alpha}_{i,k} = \alpha_{i,k} \bar{\alpha}_{i,k}^\gamma$, $\gamma$ being an optimizer parameter with suggested values in the range of $1 \leq \gamma \leq 5$. This formulation aims to reduce the influence of locally mislabeled observations by diminishing their weights exponentially based on the weights of their nearest neighbors. Incorporating this modification into the optimization process enhances robustness and promotes the identification of clearer boundaries between the models.

\section{Numerical Results}\label{sec:results}
The optimization of the loss functions for these problems was carried out using a modified Newton optimization method, as described in Appendix A. However, it is worth noting that for the flamespeed case presented in Section \ref{sec:flamespeed}, neural networks were employed and optimized using the standard Adam optimizer \cite{kingma2014adam}. All the models were implemented using Python, utilizing the Numpy \cite{van2011numpy} and Pytorch \cite{paszke2019pytorch} libraries for training and inference. The results presented in this section are based on the best case out of 5 trials.

The following subsections describe the specific application problems on which the CLSM approach was tested, highlighting its performance for each case.

\subsection{Piecewise sinusoidal function discovery}
In this section, the proposed approach is applied to an elementary test problem consisting of a piecewise function, described by Eq. \ref{eq:toyprob}. The function consists of two distinct domains - the first domain represents a sine wave with an amplitude of 0.2, while the second domain is characterized by a cosine wave with an amplitude of 0.1.

\begin{equation}\label{eq:toyprob}
f(x) = \begin{cases}
0.2 \sin(x) & \text{if } x \leq 0 \\
0.1 x \cos(x) & \text{if } x > 0
\end{cases}
\end{equation}

We generated samples for training by sampling over the domain $x \in \left[-30, 30 \right]$  at intervals of 0.05. In order to simulate real-world conditions, we add normally-distributed noise $\eta \sim \mathcal{N}(0, 0.01)$ to the function values. This results in a training dataset,  $\mathcal{D} = \{(x_i, y_i)\}_{i=1}^D$, where $y_i = f(x_i) + \eta_i$. 

Subsequently, we trained a collection of competitive models to this data, using the CLSM approach described in section \ref{sec:clsm}. During training, each "expert" model learns to specialize in a particular part of the input domain, resulting in a piecewise approximation of $f(x)$. The SE weight assigned to the MSE loss for each observation and specialized model, denoted as $\hat{\alpha}_{i,k}$, is determined based on the performance of each expert, as described by Eq. \ref{Eq:loss_function_final}.

\begin{figure}
     \centering
     \begin{subfigure}[b]{0.48\textwidth}
        \centering
        \includegraphics[width=\textwidth]{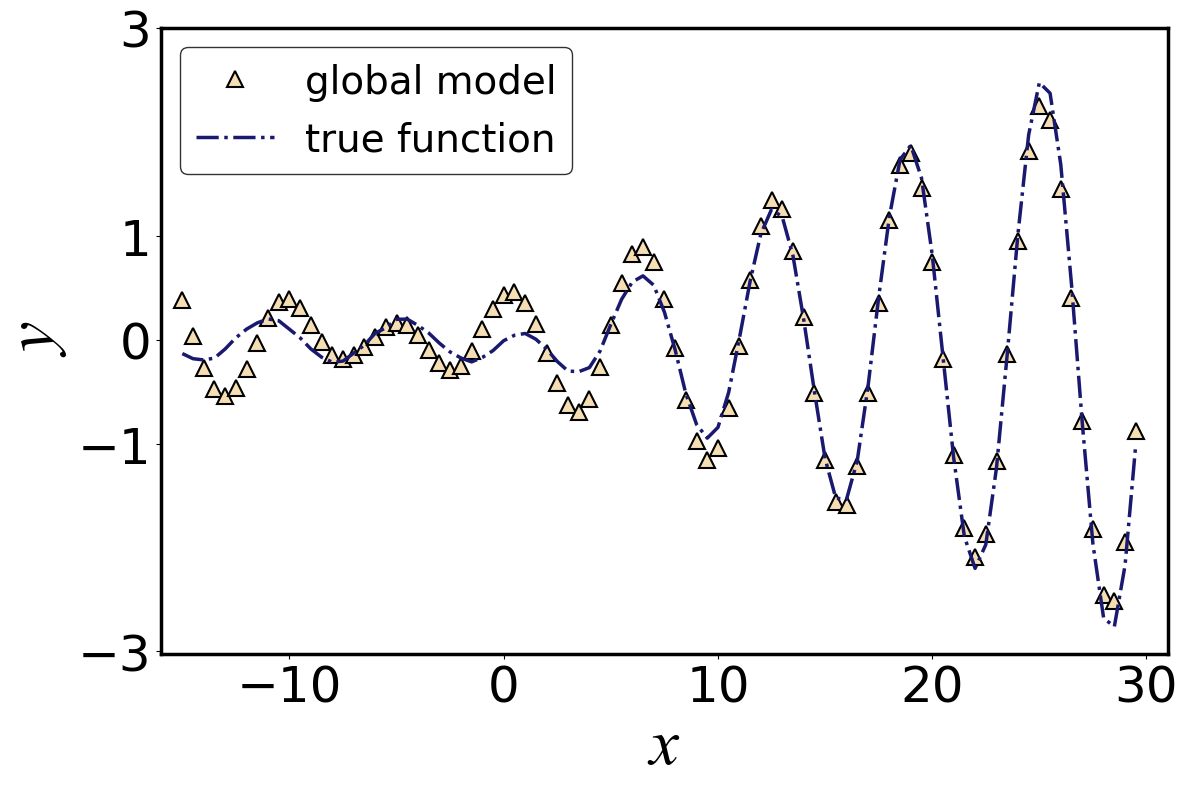}
        \caption{Global model}
        \label{fig:first_test1}
        \centering
     \end{subfigure}
     \begin{subfigure}[b]{0.48\textwidth}
        \centering
        \includegraphics[width=\textwidth]{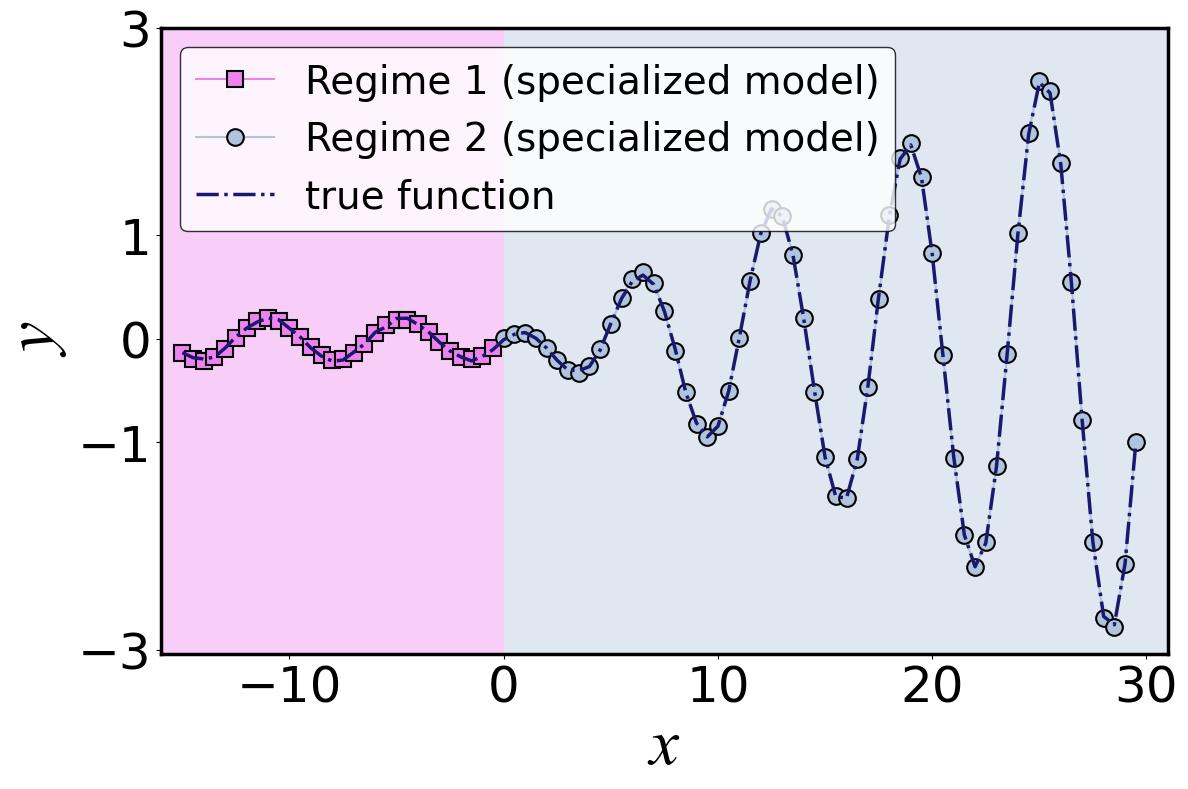}
        \caption{Specialized models}
        \label{fig:first_test2}
        \centering
     \end{subfigure}
    \caption{Figure showing a comparison of true function with (a) global model and (b) specialized models training using CLSM.}
    \label{fig:first_test}
\end{figure}

In this case, we combine LASSO regression with the CLSM approach, referred to as CLSM-LASSO. We select features that are simple functions of $\sin(x)$ and $\cos(x)$, as presented in Table \ref{tab:feature_weights_testp}. In real-life problems, domain knowledge can guide the selection of such features for function identification tasks. 

Figure \ref{fig:first_test1} compares the performance of a global model and the true function, showing significant errors, especially for $x < 0$. On the contrary, Fig. \ref{fig:first_test2} illustrates that LASSO-CLSM correctly identifies the different regimes of the function, dividing the $x$ domain at $x = 0$, leading to better agreement with the true function. Moreover, Table \ref{tab:feature_weights_testp} demonstrates that CLSM-LASSO successfully approximates the coefficients for each regime of the function. For instance, in Model 1, responsible for the regime where $x \leq 0$, the feature $\sin(x)$ has the highest coefficient, aligning with the true underlying function $0.2\sin(x)$. Similarly, in Model 2, which corresponds to the regime where $x > 0$, the feature $x\cos(x)$ has the most significant coefficient, reflecting the actual function form $0.1x\cos(x)$ in this region. Thus, the function discovered by our proposed approach can be approximated as follows:

\begin{equation}\label{eq:toyprob2}
f(x) \approx \begin{cases}
0.197 \sin(x) & \text{if } x \leq 0\\
0.1 x \cos(x) & \text{if } x > 0
\end{cases}
\end{equation}

where the final assignment for each model after training has been determined for each observation by computing $\argmax_{k}\left\{\bar{\alpha}_{i,k} \right\}_{k=1}^{Q}$. These observations demonstrate the capability of the CLSM approach to accurately identify the transition to a new function regime at $x = 0$, despite the continuous nature of the transition ($0.2 \sin(x) = 0.1 x \cos(x) = 0$ at $x = 0$). Furthermore, the CLSM-LASSO successfully approximates the true equations within each regime, showcasing its effectiveness in handling piecewise functions with distinct regimes.

\begin{table}[h!]
\centering
\setlength{\abovecaptionskip}{6pt} 
\caption{Feature-assigned weights for each specialized model}
\label{tab:feature_weights_testp}
\begin{tabular}{|c|c|c|}
\hline
\multirow{2}{4em}{\textbf{Features}} & \multicolumn{2}{c|}{\textbf{Feature coefficients}} \\
\cline{2-3}
& \textbf{Model 1} & \textbf{Model 2} \\
\hline
$x$ & $-4.5\times 10^{-5}$ & $3.1 \times 10^{-4}$ \\
\hline
sin($x$)& $6.4 \times 10^{-4}$ & 0.197\\
\hline
cos($x$)& $4.0 \times 10^{-3}$ & $1.2 \times 10^{-3}$ \\
\hline
$x$sin($x$)& $-3.9 \times 10^{-4}$ & $4.0 \times 10^{-4}$\\
\hline
sin($x$)cos($x$)& $8.8 \times  10^{-3}$ & $3.2 \times 10^{-3}$\\
\hline
$x$cos($x$)& 0.10 & $2.1\times 10^{-4}$\\
\hline
bias & $-9.9 \times 10^{-3}$ & $2.1 \times 10^{-3}$\\
\hline
\end{tabular}
\end{table}

\subsection{Damped harmonic oscillator with a time-dependent forcing}\label{sec:spring_mass1}

\begin{figure}[h]
\centering
\includegraphics[width=5cm]{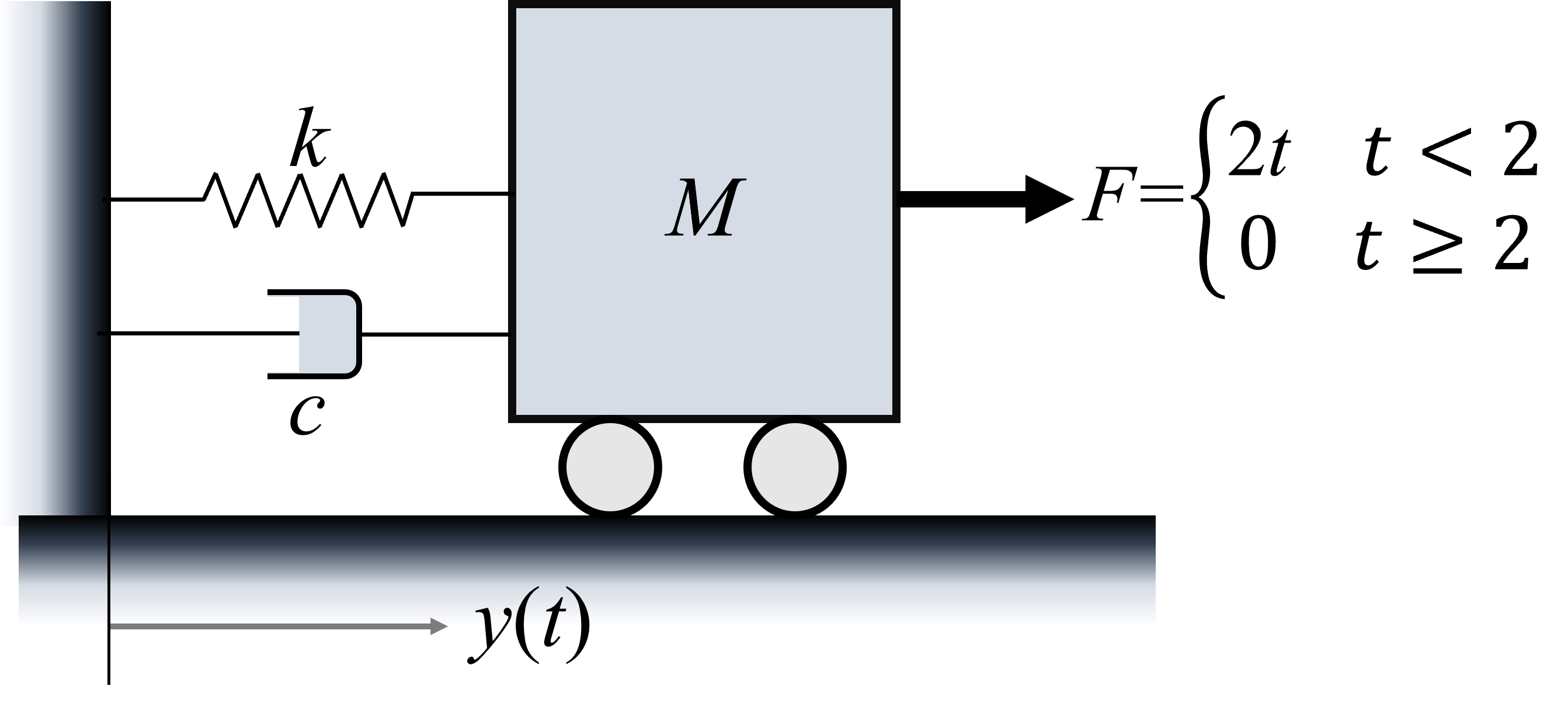}
\caption{Illustration of damped harmonic oscillator with a time-dependent forcing}
\label{fig:damped_time_dependent}
\centering
\end{figure}

Our proposed algorithm is applied to the discovery of the dynamics of a damped harmonic oscillator under the influence of a time-dependent forcing function, as shown in Figure~\ref{fig:damped_time_dependent}. Damped harmonic oscillators are widely observed in various fields of physics and engineering, and their dynamics are well-documented in the literature, making them an ideal application for our approach.

The governing equation for this damped harmonic oscillator under the influence of a time-dependent forcing function is given by:

\begin{equation}
m \ddot{y} = - c \dot{y} - k y + F(t)
\end{equation}

Where, $y$ is the displacement, $\dot{y}$ is the velocity, and $\ddot{y}$ represents the acceleration. In this equation, the parameters $m=0.75$, $c=0.05$, and $k=2.4$ represent the mass, damping coefficient, and spring constant, respectively. The term $F(t)$ denotes the external forcing function, which can be expressed as:

\begin{equation}
F(t) = 
\begin{cases} 
2t, & \text{for } t \leq 2 \\
0, & \text{otherwise} 
\end{cases}
\end{equation}

Thus, the ordinary differential equation (ODE) governing the specific system under consideration can be written as:

\begin{equation}
\ddot{y} = \begin{cases} 
- 0.0667 \dot{y} - 3.2 y, & \text{for } t \leq 2 \\
- 0.0667 \dot{y} - 3.2 y + 0.1t, & \text{otherwise}
\end{cases}
\end{equation}

To numerically solve this ODE, the time variable $t$ was discretized over the domain $t \in \left[0, 10s \right]$ using a uniform grid consisting of 500 points. The numerical solution of the ODE was obtained using the Python library Scipy \cite{virtanen2020scipy}, which provides a wrapper around the ODEPACK \cite{hindmarsh1983odepack} library of ODE solvers. The initial conditions were set as $y(0)=2$ and $\dot{y}(0)=0$. The damped oscillator being considered is driven by a linearly ramping force for a finite duration, followed by a phase of unforced oscillation with damping. This context serves as a suitable test case for evaluating our algorithm's capability to identify distinct dynamical regimes within such systems.

To obtain specialized models in this case, we integrate CLSM with the approach for discovering dynamical systems, SINDy \cite{brunton2016discovering}, and refer to this method as CLSM-SINDy. First, observations of the oscillator at discrete time points are collected to form a dataset $\mathcal{X} = \left\{\bm{y}_i \right\}_{i=1}^T$, where $\bm y_i = \left[y_i, \dot{y}_i \right]$ represents the state of the damped oscillator at the $i-$th observation. From these observations, a library of features, $\bm \Theta(\bm y_i)$, is constructed. This library consists of polynomial functions of $y$ and $\dot{y}$ arranged in increasing order (such as $y, \dot{y}, y\dot{y}, y^2, \ldots$). These features serve as inputs for LASSO regression in all specialized models, with the objective of discovering the true governing equations, similar to the SINDy approach \cite{brunton2016discovering}. The integrated CLSM-SINDy method combines the competitive learning process of CLSM with the model discovery capabilities of SINDy, enabling the identification and characterization of the underlying dynamics of the damped oscillator system.

\begin{figure}
     \centering
     \begin{subfigure}[b]{0.42\textwidth}
        \centering
        \includegraphics[width=\textwidth]{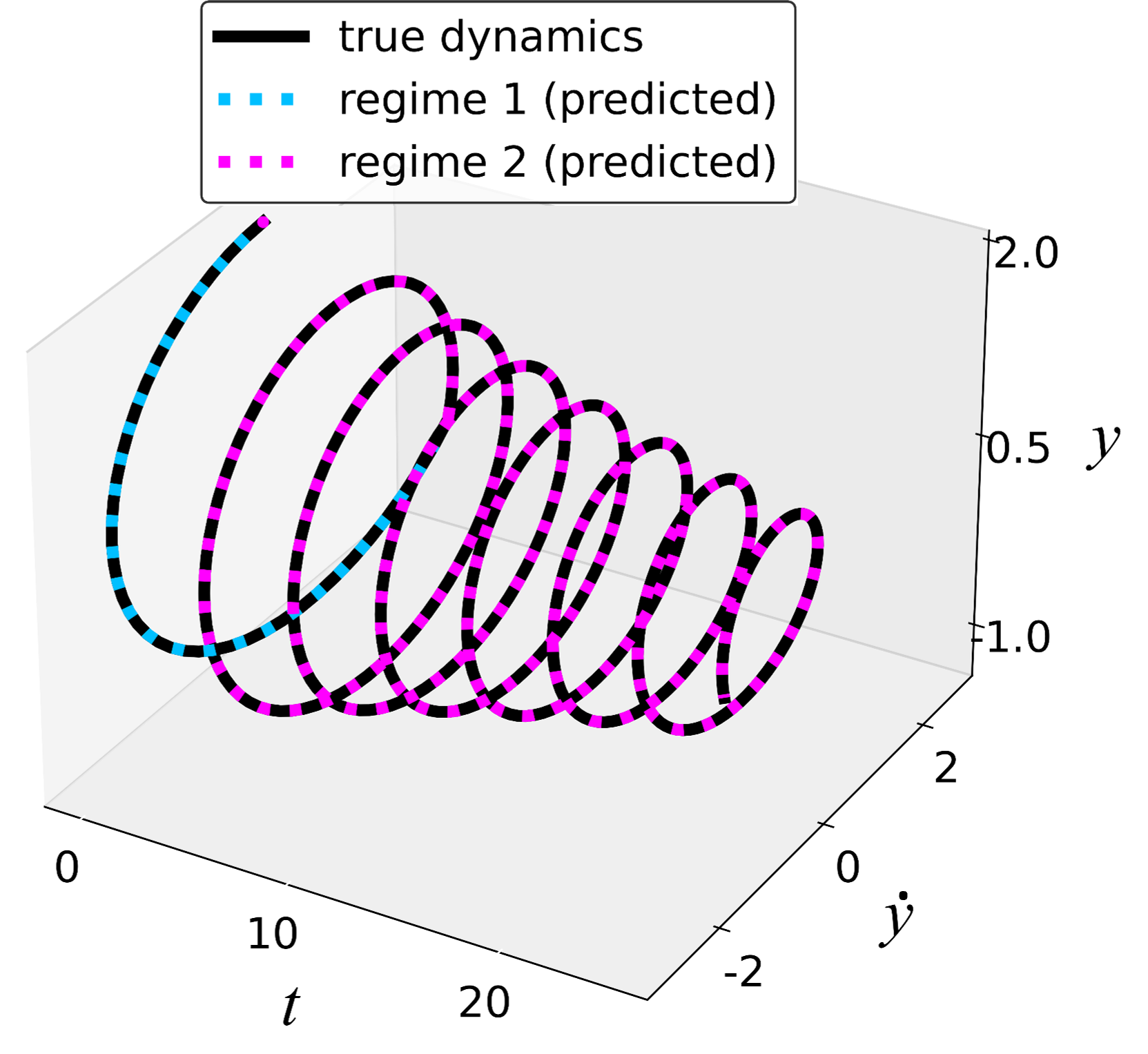}
        \caption{}
        \label{fig:mass_damper1a}
        \centering
     \end{subfigure}
     \hfill
     \begin{subfigure}[b]{0.50\textwidth}
        \centering
        \includegraphics[width=\textwidth]{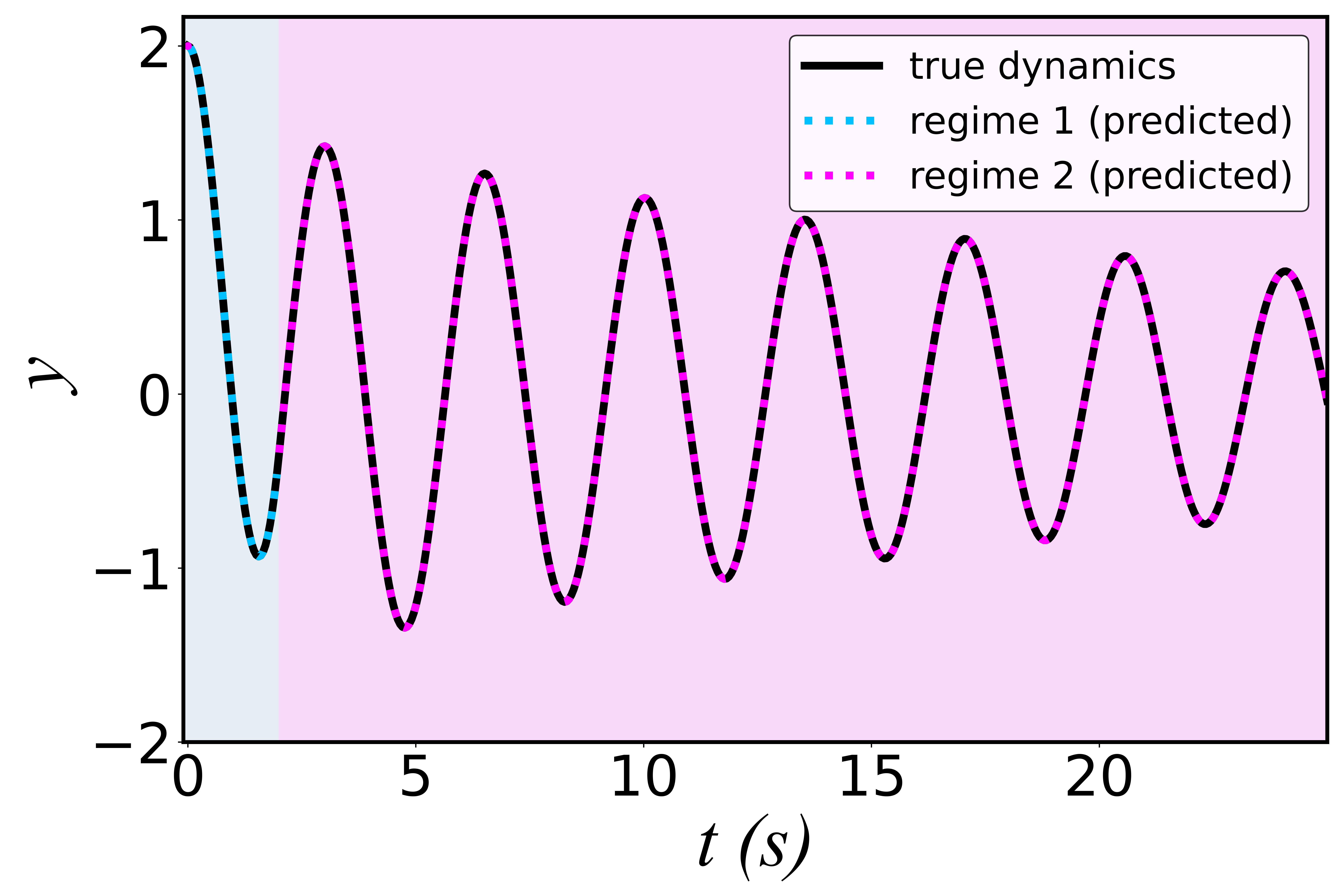}
        \caption{}
        \label{fig:mass_damper1b}
        \centering
     \end{subfigure}
    \caption{Comparison between exact and learned dynamics for harmonic oscillator with a time-dependent forcing}
    \label{fig:mass_damper1}
\end{figure}

The CLSM-SINDy approach to discovering specialized models, as depicted in the 3-dimensional plot of the system dynamics in Fig. \ref{fig:mass_damper1a}, successfully captures the true dynamics of the system. Furthermore, as shown in Fig. \ref{fig:mass_damper1b}, which illustrates the displacement of the oscillator over time, the proposed approach accurately identifies the switch to an unforced state at $t=2$. The discovered equation by CLSM-SINDy for this problem is:

\begin{equation}\label{eq:CLSM_spring_mass1}
\ddot{y} \approx \begin{cases}
-0.06665 \dot{y} - 3.2000 y, & \text{for } t \leq 2 \\
- 0.0667 \dot{y} -3.19160 y + 2.00359 t, & \text{otherwise}
\end{cases}
\end{equation}

Note that in Eq. \ref{eq:CLSM_spring_mass1}, the other features, which have coefficients with absolute values less than $3 \times 10^{-3}$, have been disregarded. These results demonstrate the effectiveness of the CLSM approach in learning the dynamics of the damped oscillator with time-dependent regimes.

\subsection{Damped harmonic oscillator with a displacement-dependent damping and restoring force}
\begin{figure}[h]
\centering
\includegraphics[width=5cm]{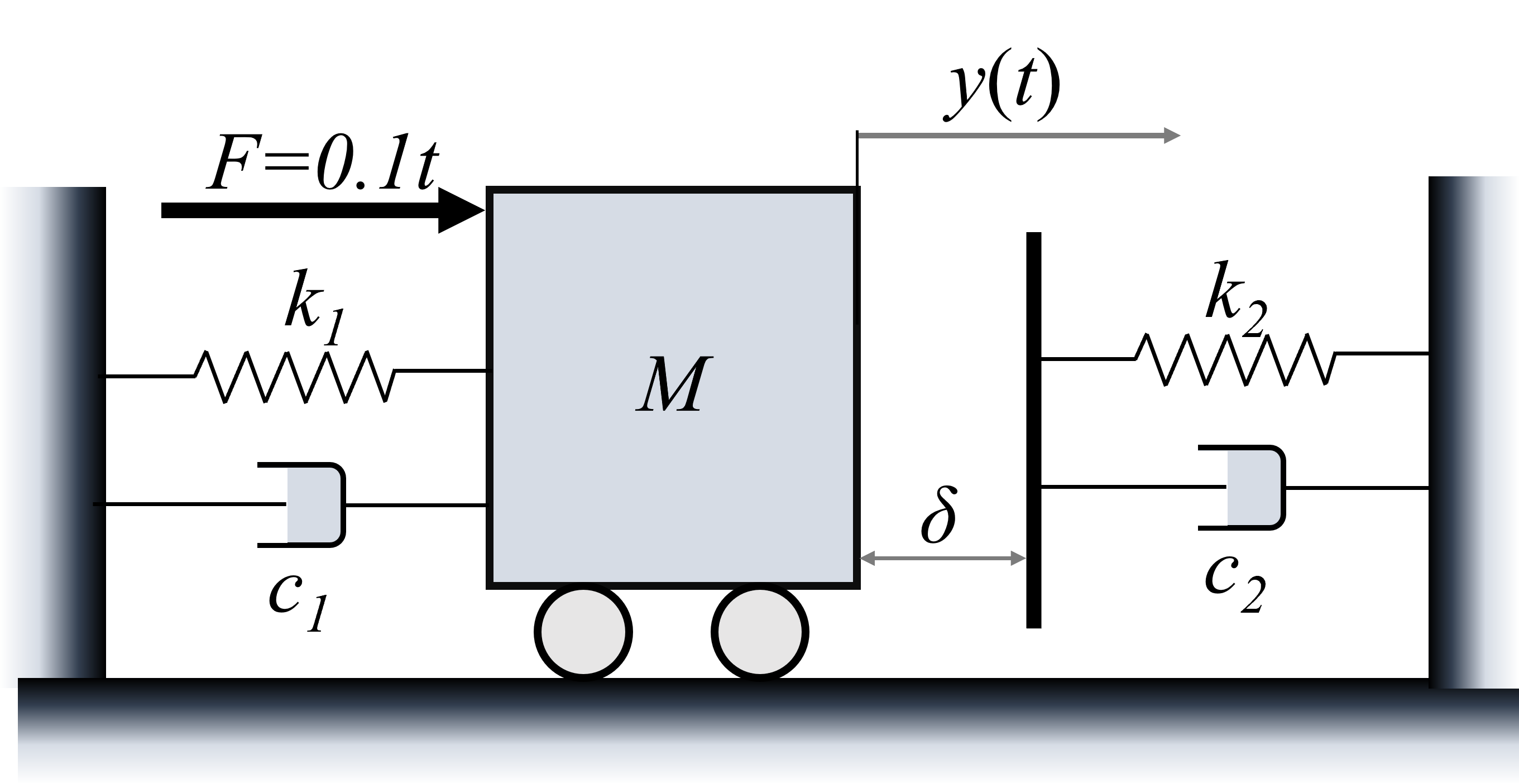}
\caption{Illustration of damped harmonic oscillator with a displacement-dependent damping and restoring force.}
\label{fig:spring_mass_displacement}
\centering
\end{figure}

Next, we present the results for a damped harmonic oscillator with an additional nonlinear component, where an extra damper and spring are activated based on the position of the mass. This system poses a  challenging test case due to the presence of additional components that are conditionally applied depending on the displacement of the oscillator. Figure \ref{fig:spring_mass_displacement} provides a visual representation of such a system. The governing equation for this system is given by:

\begin{equation}
\ddot{y} = -\frac{{c_1}}{{m}} \dot{y} - \frac{{k_1}}{{m}}y + 0.1t - \left[\frac{c_2}{m}\ddot{y} + \frac{k_2}{m}(y - \delta) \right] U(y - \delta)
\end{equation}
where $U(y - \delta)$ is a step function, defined as
\[
U(y - \delta) = \begin{cases}
0 & \text{if } y \leq \delta \\
1 & \text{if } y > \delta \\
\end{cases}
\]

In the equation, \(m\) represents the mass of the oscillator, \(c_1\) and \(k_1\) denote the damping coefficient and the spring constant for the primary oscillator respectively. \(F(t)\) is the time-dependent external forcing term, and \(\delta\) is a displacement threshold. When the displacement exceeds the threshold (\(y > \delta\)), the oscillator activates additional damping and restoring forces characterized by \(c_2\) and \(k_2\), respectively.
The goal is to accurately discover the underlying dynamics and identify the conditions under which the additional forces come into play.

The values used for the system parameters are as follows: $m=1.0$, \(c_1 = 0.25\), \(k_10\), \(c_2 = 0.15\), \(k_2 = 5\), \(\delta = 0.25\). The system is initialized by displacing it to $y_0 = 1$ with $\dot{y} = 0$. An external force that is a linear function of time, $F\left(t \right) = 0.1t$, is applied. the governing equation for this specific case, given the parameters listed, is:

\begin{equation}
\ddot{y} = \begin{cases}
-0.25 \dot{y} - 10 y + 0.1t & \text{if } y \leq \delta \\
-0.40 \dot{y} - 15 y + 0.1t + 1.25 & \text{if } y > \delta \\
\end{cases}
\end{equation}

\begin{figure}
     \centering
     \begin{subfigure}[b]{0.42\textwidth}
        \centering
        \includegraphics[width=\textwidth]{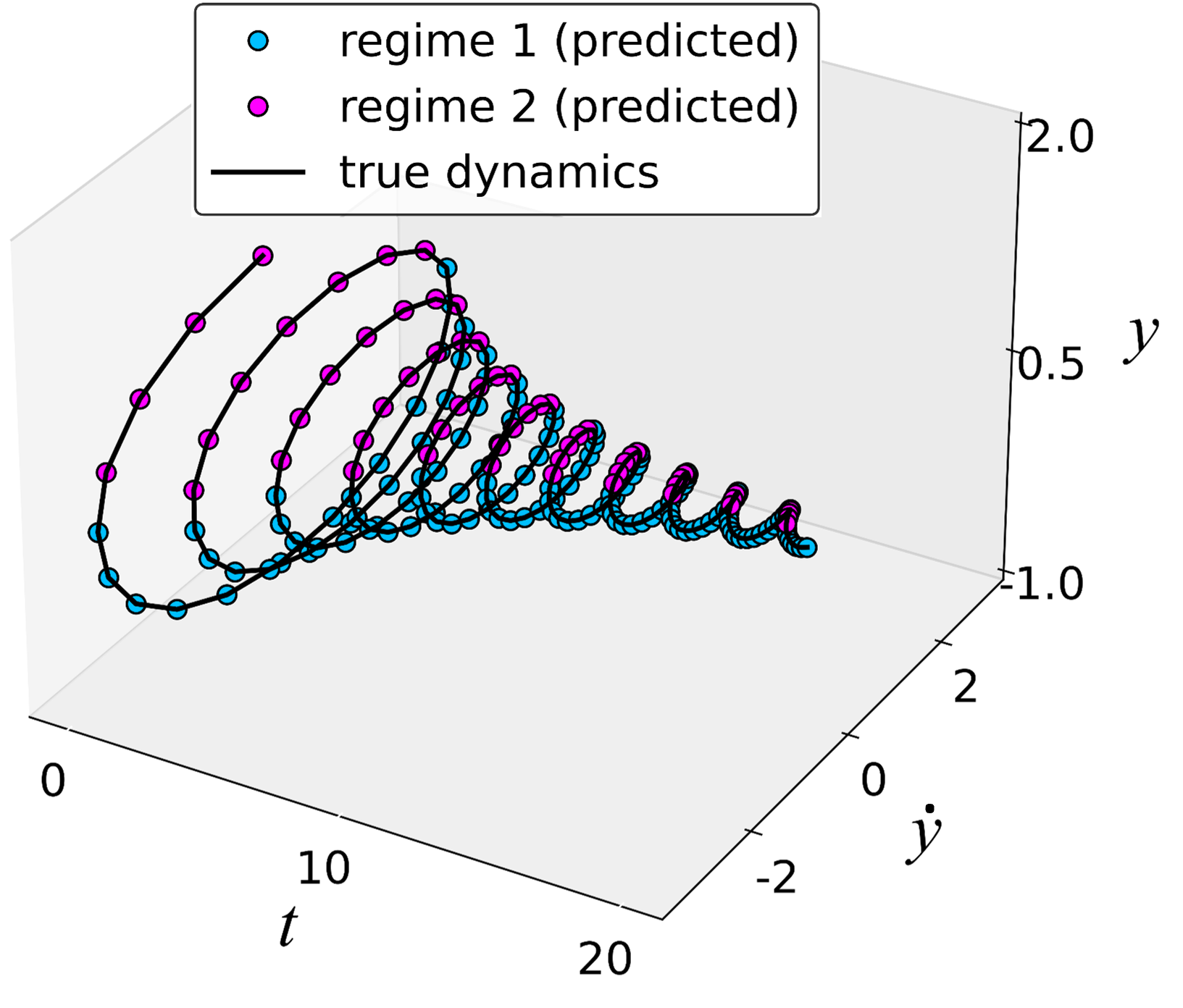}
        \caption{scatter plot of $y$ against $x$}
        \label{fig:spring_mass_3D}
        \centering
     \end{subfigure}
     \hfill
     \begin{subfigure}[b]{0.50\textwidth}
        \centering
        \includegraphics[width=\textwidth]{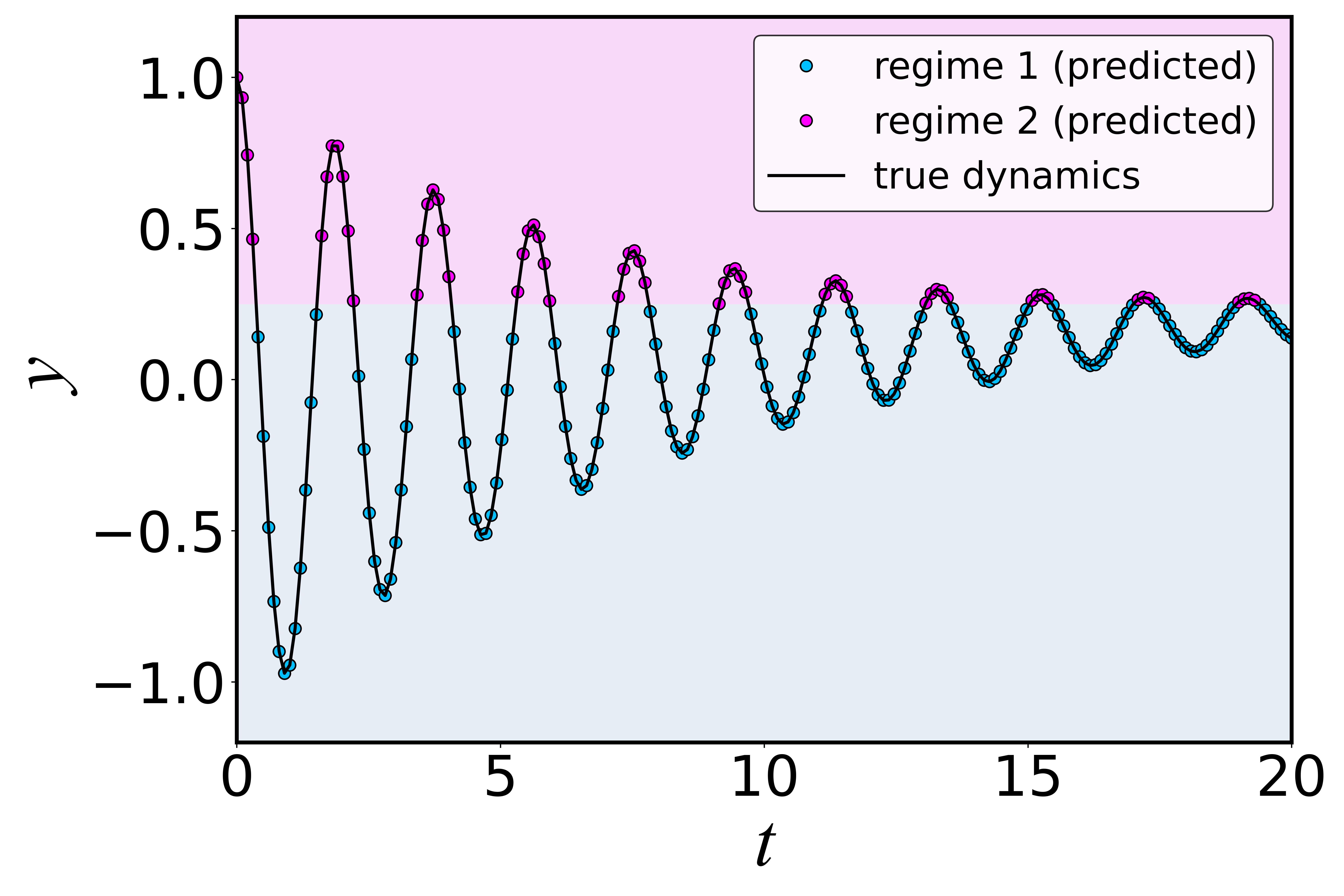}
        \caption{loss landscape with all training samples}
        \label{fig:spring_mass_2D}
        \centering
     \end{subfigure}
    \caption{Comparison between exact and learned dynamics for harmonic oscillator with displacement-dependent damping and restoring force}
    \label{fig:spring_mass_im}
\end{figure}

Similar to the previous case in section \ref{sec:spring_mass1}, the time variable $t$ was discretized over the domain $t \in \left[0, 10s \right]$ using a uniform grid of 200 points. After solving the ODE to obtain training data, polynomial features of the state observations are collected to form a library. Then, a set of specialized models are competitively trained on the data, using the CLSM-SINDy approach. The resulting equation discovered is:

\begin{equation}
\ddot{y} = \begin{cases}
-0.25 \dot{y} - 10 y + 0.1t & \text{if } y \leq \delta \\
-0.40 \dot{y} - 15 y + 0.101t + 1.23 & \text{if } y > \delta \\
\end{cases}
\end{equation}

which approximates the true model. The dynamical behavior of the true and learned systems is shown in Fig. \ref{fig:spring_mass_3D}, showing that the learned dynamics closely approximate the true dynamical system. Furthermore, Fig. \ref{fig:spring_mass_2D} more clearly shows that, even though the switching point appears smooth and difficult to spot visually, the learned model correctly identifies $y \approx 0.25$ as the regime transition point.

The results underscore the CLSM-SINDy's ability to accurately identify and separate different regimes within a system and determine the governing equations within each regime, even in the presence of nonlinearity and time-dependent forces. 

\subsection{Laminar flame speed correlation} \label{sec:flamespeed}

The previous applications demonstrate the effectiveness of CLSM in model discovery using feature engineering and selection through regularization. However, our methodology extends beyond these specific techniques and can be applied to a wide range of regression models, including neural networks, ridge regression, and others. This flexibility allows our approach to be utilized in various applications where gradient-based optimization is employed for learning, even when the model form is not interpretable as in LASSO. Although the interpretability of the specialized models and their corresponding regimes may be challenging in such cases, CLSM still provides lower training and test errors or reduced model complexity.

In order to demonstrate the broad applicability of CLSM, we extend its application to another problem, namely, flame propagation in premixed combustion scenarios, where empirical models of laminar flame speed are often coupled with computational fluid dynamics (CFD) models to simulate turbulent flame behavior.  The speed at which a flame front propagates through a combustible mixture, known as the flame speed ($s_L$), is influenced by various parameters such as fuel-air ratios, temperature, pressure, and intensity of turbulence in the combustion chamber. Traditional empirical models based on simplified correlations struggle to capture the complexity of real-world combustion processes, particularly when there are multiple regimes of flame propagation behavior.

\begin{figure}[!h]
\centering\includegraphics[width=3.5in]{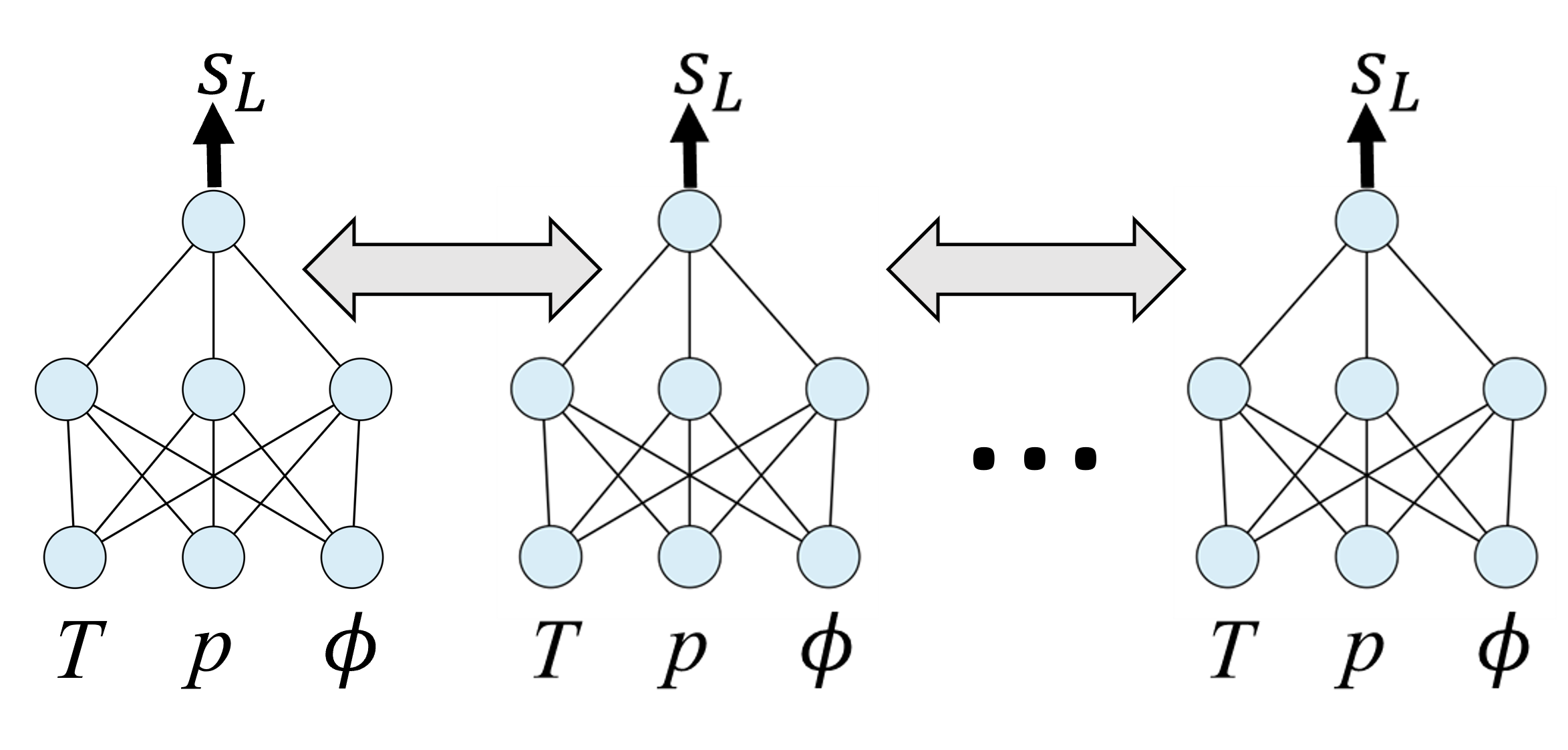}
\caption{Illustration of CLSM-NN approach used for training flame speed data. Each model receives an input of p, T, and $\phi$, and predicts the flame speed, $s_L$. The double-headed arrows represent the competitive training process between the models.}
\label{fig:flamespeedNNs}
\end{figure}

Our proposed model tackles these challenges by simultaneously identifying and partitioning the different regimes within the input space and fitting accurate flame speed correlations specific to each regime. The model is trained on a dataset of flame speed numerical simulations, gathered from a range of thermodynamic conditions and fuel-air ratios. The dataset was generated using a chemical kinetics software, Cantera \cite{cantera} to simulate flame propagation under varying conditions of equivalence ratio, pressure, and temperature using a methane-air combustion mechanism \cite{smith1999gri}. 4150 samples were generated, from which 40 \% of the points were reserved for testing. 

For this problem, the specialized models are neural networks (NNs), as illustrated in Fig. \ref{fig:flamespeedNNs}, resulting in a CLSM-NN approach. These neural networks take three key operating variables: pressure ($p$), temperature ($T$) and equivalence ratio ($\phi$) as inputs and predict flame speeds. In our experiments, we varied the number of specialized models and obtained the best corresponding test MSEs based on 5 trials. The results indicate that the global model achieved a test MSE of $5.65 \times 10^{-4}$ and $R^2 = 0.9810$, while the use of two specialized models reduced the test MSE to $4.65 \times 10^{-5}$ and resulted in an improved $R^2$ of $0.9983$. Increasing the number of specialized models to three results in a less significant performance improvement, with a test MSE of $4.08 \times 10^{-5}$ and $R^2 = 0.9984$. These results highlight the effectiveness of CLSM-NN in capturing the complexities and variations within the flame-speed data, thereby providing more accurate predictions.

\begin{figure}
     \centering
     \begin{subfigure}[b]{0.3\textwidth}
        \centering
        \includegraphics[width=\textwidth]{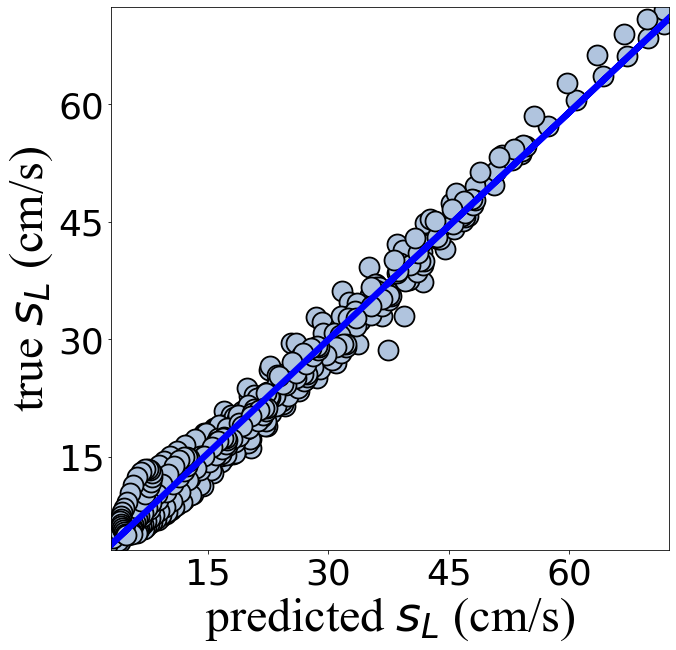}
        \caption{global model: $R^2 = 0.9810$}
        \label{fig:par1}
        \centering
     \end{subfigure}
     \hfill
          \begin{subfigure}[b]{0.3\textwidth}
        \centering
        \includegraphics[width=\textwidth]{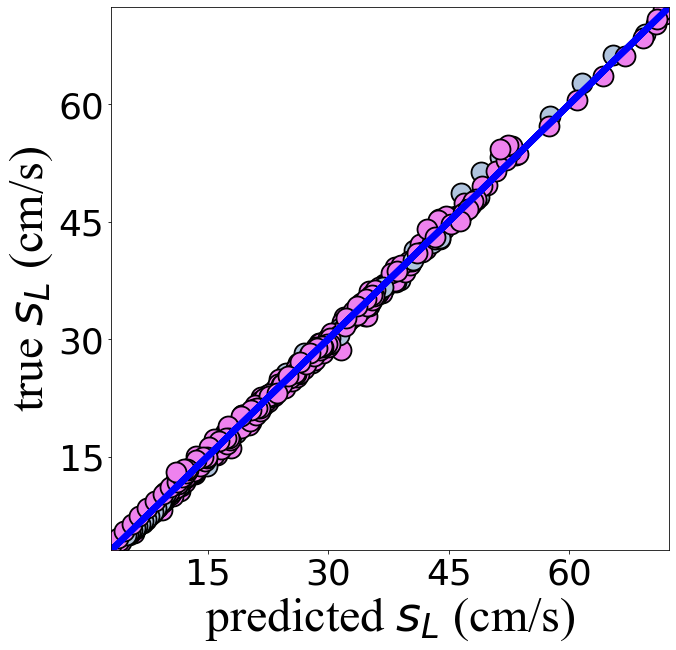}
        \caption{2 specialized models: $R^2 = 0.9983$}
        \label{fig:par2}
        \centering
     \end{subfigure}
     \hfill
     \begin{subfigure}[b]{0.3\textwidth}
        \centering
        \includegraphics[width=\textwidth]{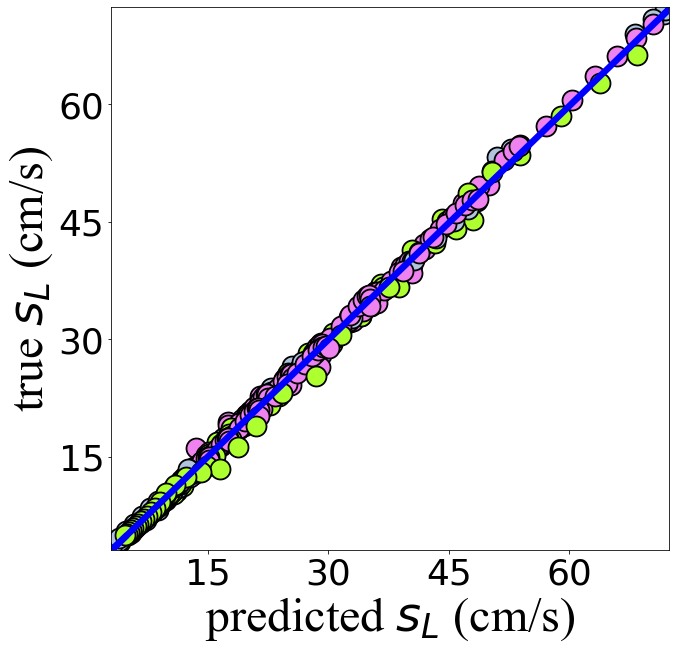}
        \caption{3 specialized models: $R^2 = 0.9984$}
        \label{fig:par3}
        \centering
     \end{subfigure}
    \caption{Parity plots for various numbers of specialized models. Different colors are used to represent the assignment to various specialized models.}
    \label{fig:parity_plots}
\end{figure}
\begin{figure}
     \centering
     \begin{subfigure}[b]{0.45\textwidth}
        \centering
        \includegraphics[width=\textwidth]{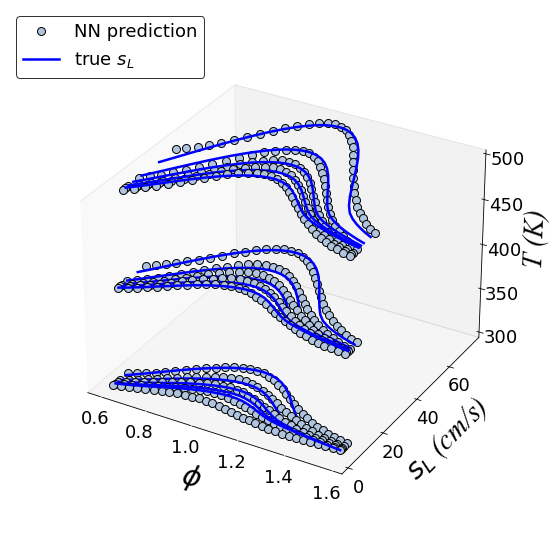}
        \caption{Global model}
        \label{fig:flamespeed1}
        \centering
     \end{subfigure}
     \hfill
          \begin{subfigure}[b]{0.45\textwidth}
        \centering
        \includegraphics[width=\textwidth]{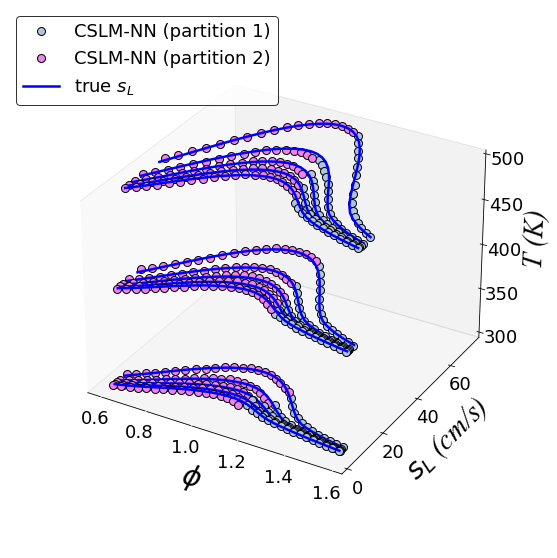}
        \caption{2 specialized models}
        \label{fig:flamespeed2}
        \centering
     \end{subfigure}
    \caption{Laminar flame speed comparison for global and two-model CLSM-NN. The multiple curves on the same $\phi$-$s_L$ plane represent a different pressure level, with higher pressures corresponding to higher flame speeds.}
    \label{fig:flamespeed}
\end{figure}

Fig. \ref{fig:parity_plots} presents parity plots based on the test data, illustrating the correlation between predicted and actual flame speed values for the three different cases: the global model, the two-model approach, and the three-model approach. These plots reveal that as the models become more specialized, the test performance improves, as indicated by higher $R^2$ values. Fig. \ref{fig:flamespeed} shows plots of the flame speed predictions and assignment of observations to various partitions, for both global and two-model CLSM-NN cases. The global model results in significant errors, especially at lower values of temperature. In contrast, the 2-model case splits the problem between two neural networks, resulting in better agreement with the data. Furthermore, it can be observed that CLSM-NN approach splits the input space close to a region of physical significance, i.e., close to stoichiometric values of $\phi$ where we have just enough oxidizer to burn the fuel. Thus, in this case, it can be seen that in addition to improving test accuracy over a global model, the CLSM-NN approach also leads to partitions that are physically meaningful. The CLSM-NN approach does not make any significant distinctions based on pressure or temperature, suggesting that flame speed exhibits the most complexity in its dependence on $\phi$.

\section{Conclusion}
In this paper, we have introduced the Competitive Learning of Specialized Models (CLSM) approach, which involves training a set of specialized models on a dataset. The CLSM approach modifies the loss function during training to weigh poorly learned observations less and emphasize well-learned observations more. We have demonstrated the effectiveness of CLSM in learning complex systems with multiple regimes, overcoming the limitations of traditional global models that struggle to capture the distinct behavior of different regimes.

The CLSM was combined with various regression approaches and applied to various applications, including model discovery of a synthetic problem, dynamical system discovery of two harmonic oscillators with piecewise behavior, and estimating the laminar flame speed of a methane-air flame at various temperatures, pressure, and equivalence ratios. For both the synthetic problem and harmonic oscillators, the CLSM approach successfully identified the locations of regime transitions and reproduced the correct model coefficients. In the flame speed application, increasing specialization led to better test errors and the partitions identified by the model were physically consistent.

However, like all models, the proposed CLSM approach is not without limitations. A primary difficulty lies in determining the number of specialized models to employ. Incremental testing, starting with a small number of models and gradually increasing specialization until the model errors plateau, can help in this regard. Also, the CLSM approach may occasionally converge to a local minima during training. The modified squared error weights that consider the weights of neighboring points (as presented in Eq. \ref{sec:clsm}) mitigates this issue, but multiple trials may still be necessary to reach the best regime partitions.

In conclusion, the presented CLSM approach represents a significant advancement in modeling complex systems. Its ability to identify and partition different behavior regimes, as well as generate accurate correlations within each regime, enables more detailed and precise predictions. This approach has broad applicability and holds promise for various domains, where understanding and modeling complex systems are crucial for improved analysis, design, and optimization.

\appendix
\setcounter{secnumdepth}{0}
\section{Appendix A. Modified Newton Method used for Optimization}
We present a mathematical description of a modified Newton method for optimization, which is used to optimize the models in section \ref{sec:results}. The Newton method is a second-order optimizer that utilizes the Hessian matrix and first-order gradients to update the parameters of a model. In our case, we consider a general scenario with trainable parameters represented as a vector, $\bm \theta = \left[\theta_1, \theta_2, ..., \theta_P \right]$.

In the standard Newton method, the parameters at iteration $k + 1$ are obtained by updating the parameters at iteration $k$ as:

\begin{equation}
\bm \theta^{k+1} = \bm \theta^{k} - \bm {\Delta \theta},
\end{equation}

where $\bm {\Delta \theta} = \bm {H}_k^{-1} \bm{g}_k = \left[{\Delta \theta}_1, {\Delta \theta}_2, ..., {\Delta \theta}_P \right] \in \mathbb{R}^P$ is the change in the parameter vector, $\bm{g}_k$ is the gradient of the loss function with respect to the parameters: $\bm g_k = \left[\partial \mathcal{L}/\partial \theta_1, \partial \mathcal{L}/\partial \theta_2, ..., \partial \mathcal{L}/\partial \theta_P \right]^T \in \mathbb{R}^{P \times 1}$. In this study, to promote robustness and reduce the propensity of the proposed approach to converge to local minima, we have made two changes to the standard Newton optimization method by updating the parameters according to the following rule: 

\begin{equation}
\bm {\Delta \theta} = \textrm{clip} \left[ \left(\bm{H}_k + \bm{\mathcal{E}} \right)^{-1} \bm{g}_k, \beta \right]
\end{equation}

In the above equation, we add a diagonal matrix $\bm{\mathcal{E}} = \textrm{diag}\left(\bm \epsilon\right) \in \mathbb{R}^{P \times P}$ to the original Hessian matrix. The vector $\bm \epsilon \in \mathbb{R}^P$ contains elements sampled from a uniform distribution, such that $\bm \epsilon \sim \mathcal{U} \left(0, \epsilon_{\textrm{max}} \right)$, where $\epsilon_{\textrm{max}}$ represents the maximum desired noise level. By introducing some noise to the diagonal of the Hessian matrix, we introduce variability into the step direction at each iteration and prevent matrix inversion from becoming singular. Before updating the current parameters, $\bm \theta^k$, the product of the modified Hessian and the gradient, $\bm g_k$, is clipped such that $ \max \left| \bm{\Delta \theta} \right|\leq \beta$. This clipping ensures that the maximum absolute change in the parameters remains within the specified limit. As all models learn concurrently and the loss landscape evolves during the training process, the clipping mechanism slows down the learning process and prevents premature convergence based on the current loss landscape.

\bibliographystyle{unsrtnat}
\bibliography{references}  

\end{document}